# Bridgeout: stochastic bridge regularization for deep neural networks

Najeeb Khan
Department of Computer Science
University of Saskatchewan, Canada.
najeeb.khan@usask.ca

Jawad Shah
British Malaysian Institute
Universiti Kuala Lumpur, Malaysia.
jawad@unikl.edu.my

Ian Stavness[✉]
Department of Computer Science
University of Saskatchewan, Canada.
ian.stavness@usask.ca

## Abstract

*A major challenge in training deep neural networks is* overfitting, *i.e. inferior performance on unseen test examples compared to performance on training examples. To reduce overfitting, stochastic regularization methods have shown superior performance compared to deterministic weight penalties on a number of image recognition tasks. Stochastic methods such as Dropout and Shakeout, in expectation, are equivalent to imposing a ridge and elastic-net penalty on the model parameters, respectively. However, the choice of the norm of weight penalty is problem dependent and is not restricted to $\{L_1, L_2\}$. Therefore, in this paper we propose the Bridgeout stochastic regularization technique and prove that it is equivalent to an $L_q$ penalty on the weights, where the norm q can be learned as a hyperparameter from data. Experimental results show that Bridgeout results in sparse model weights, improved gradients and superior classification performance compared to Dropout and Shakeout on synthetic and real datasets.*

## 1. Introduction

Deep neural networks (DNN) are expressive machine learning models that have been effective on many difficult computer vision tasks involving large amounts of image data. Being supervised machine learning models, DNNs are trained by minimizing the discrepancy between the model output and the original labels of the images in a training set. The goal, however, is to minimize the error in labeling previously unseen data known as the generalization error. Thus, training DNNs is an optimization problem where the training error serves as a proxy for the true objective: the generalization error [5]. When the complexity of the model is roughly the same as that of the task, the training error serves as a faithful proxy for the generalization error. However, with the expressive power of DNNs, even small architectures can capture the random noise in the training samples and therefore result in high generalization error.

To overcome this problem, researchers have devised different strategies to prevent DNNs from misinterpreting random variations in the training data as patterns responsible for the labels. Increasing the training dataset size is one potential solution, but often not possible. Augmenting the data with new samples that are slight variations of the original samples is also a commonly used approach. Early-stopping, i.e. stopping the training process before the validation error starts ascending, is another effective way to stop overfitting. While early-stopping is the easiest to exercise, in practice it does not match the performance achieved by more sophisticated techniques that regularize the models [10].

The *simplest* model that fits the training data will generalize better than more *complex* models. There is, however, no easy way to choose a simple model that will yield the best performance, and a simple model may perform worse due to sensitivity to initial conditions and the bias–variance tradeoff [27]. Therefore, a common approach is to start with a large neural network and then constrain the model in some way to prevent it from learning sampling noise. This process is known as regularization. Deterministic techniques either prune the network by removing less important neurons or impose a weight penalty on the magnitude of the weights of each layer. Penalizing the weights with the $L_1$ norm can be seen as feature selection procedure, whereas, penalizing the $L_2$ norm of the weights can be interpreted as continuous shrinkage of the weights, which prevents overly complicated decision boundaries.

Stochastic regularization techniques approach overfitting by constructing an ensemble of poorly trained models and then averaging their predictions. Given a neural network, for each training example in the training set, Dropout [30] sets the units in the network to zero with a probability $1-p$. Thus each training example is used to train a slightly different network. At inference time, all the units in the network are kept but their outputs are scaled with $p$ which serves as averaging the prediction of many networks. Dropout is equivalent to a ridge penalty on the model weights ($L_2$ norm). Shakeout [15], a technique similar to Dropout, where all the outgoing weights from a unit are either set to a signed constant or incremented by a signed constant. Shakeout can be interpreted in terms of deterministic regularization techniques as performing both ridge and lasso ($L_2$ and $L_1$ norm) regularization.

Current stochastic methods implicitly result in a weight penalty whose norm is decided *a priori* independent of the dataset. Since different datasets may require different norm of the weight penalty [7], we hypothesize that a stochastic method with an adaptive norm will result in superior performance to fixed-norm stochastic methods, such as Dropout and Shakeout. Also, an adaptive norm formulation is more general, and therefore would incorporate the fixed norm methods as special cases.

In this work, we propose Bridgeout: stochastic regularization with an adaptive norm. We theoretically prove that Bridgeout is equivalent to $L_q$ weight penalty for the generalized linear models (GLM) and show that for $q = 2$ it is equivalent to Dropout. We empirically verify our theoretical results for DNNs and show that Bridgeout results in better image classification performance than Dropout and Shakeout on MNIST and Fashion-MNIST datasets.

The rest of the paper is organized as follows: Section 2 provides the background and works related to our main contribution, followed by a description of the Bridgeout stochastic regularization in Section 3. Section 4 describes experimental results. Discussion and summary are given in Sections 6 and 7.

## 2. Background and Related Work

### 2.1. Feedforward Neural Networks

In this paper we propose a regularization method for fully connected feedforward neural networks. Consider a neural network with $L$ layers, the output of the $l$-th layer with weights $\boldsymbol{W}^l \in \mathcal{R}^{k \times d}$ is given by

$$\boldsymbol{\nu}^l = \boldsymbol{W}^l \boldsymbol{a}^{l-1} + \boldsymbol{b}^l, \tag{1}$$

$$\boldsymbol{a}^l = \sigma(\boldsymbol{\nu}^l), \tag{2}$$

for $l = 1 \cdots L$, where $\boldsymbol{a}^{l-1}$ is the output of layer $l-1$, $\boldsymbol{b}^l$ is a bias vector, $\sigma$ is a non-linear activation function and $\boldsymbol{a}^0$ is the input to the network. The weights of the neural network are trained by minimizing a cost function $J$ such as cross entropy, over the training set. The minimization is done using variants of the gradient descent algorithm. The gradients of the cost function with respect to network weights are calculated using the backpropagation algorithm [18]. The $i$-th update of weights of the $l$-th layer is as follows

$$\boldsymbol{W}_{i+1}^l = \boldsymbol{W}_i^l - \mu \frac{\partial J}{\partial \boldsymbol{W}_i^l}, \tag{3}$$

where $\mu$ is the learning rate.

### 2.2. Deterministic Regularization

Deterministic regularization methods constrain the neural network model directly based on the model structure and the training data. Pruning and weight penalties are the two dominant deterministic regularization techniques used in neural networks.

#### 2.2.1 Pruning

Pruning attempts to match the size of the model to that which is inherently required for the problem by removing redundant neurons from the network. A number of different pruning methods have been proposed to identify the redundant neurons (see Reed [27] for review), including skeletonization based on error gradient [22] and optimal brain damage based on the Hessian of the error with respect to a particular neuron [20]. Recently Han et al. [12] proposed magnitude based pruning, which permanently drops connections that have low magnitude weights followed by re-training the pruned network. The authors reported significant reductions in computations and memory usage on state of the art image classification tasks without affecting classification accuracy.

#### 2.2.2 Weight Penalties

While pruning explicitly removes redundant parts of the network, weight penalization methods add a penalty term to the cost function, so as to favour simpler models over more complicated ones, in terms of weight magnitudes, during training.

The most popular weight penalization method is the ridge regularization, which adds the $L_2$ norm of the network weights to the cost function [14]. Ridge regularization continuously shrinks the network weights during training. While ridge regularization achieves smaller weights and better generalization error, it does not result in a sparse weight matrix of the trained network, which indicates that ridge regression is useful when all the input features are important.

Sparse weights are desirable in networks for problems where some input features are unimportant or noisy. This is often the case in high dimensional problems such as image classification where, although, the images are high dimensional, images belonging to the same class exhibit *degenerate structure*, lying near a low-dimensional manifold [34]. Sparse models can exploit such low-dimensional structure. Therefore, lasso regularization has also been previously proposed [31], which adds the $L_1$ norm of the model weights to the cost function. Elastic-net regularization has also been proposed to combine both $L_1$ and $L_2$ norms of the weights [39].

Towards a more general form of regularization, Frank and Friedman [7] proposed to optimize for the norm of the weight penalty based on the problem at hand, known as *bridge* regularization. It has been shown that bridge regularization performs better than ridge, lasso and elastic-net in certain regression problems [26]. Besides linear regression, bridge regularization has been applied to support vector machines [21] with strong results. As a special case of bridge regularization, $L_{1/2}$ has been shown to exhibit useful statistical properties including sparseness and unbiasedness [36]. Different training algorithms have been proposed for training neural networks with $L_{1/2}$ weight penalty [6, 37].

In terms of Bayesian estimation, ridge and lasso penalties imply a Gaussian and Laplacian prior on model weights, respectively. On the other hand, an $L_q$ penalty corresponds to the Generalized Gaussian prior on the model weights [7]. Generalized Gaussian distribution [23] is more comprehensive encompassing Gaussian and Laplacian distributions as special cases.

### 2.3. Stochastic Regularization

In contrast to deterministic methods that only depend on the training data set and the network weights, stochastic methods add random noise to the model. Adding random noise to the model reduces the correlation between the neural activations, which result in robust performance and better generalization. Different theoretical interpretations for stochastic regularization methods have been proposed, including their equivalence to the deterministic methods when the randomness is marginalized and as an approximation to Bayesian model averaging. In practice, stochastic methods have shown superior performance to that of deterministic regularization methods in a wide range of problems [4, 17].

#### 2.3.1 Dropout

Dropout [30] randomly drops units from the network during training with probability $1-p$. For each training example, a random binary mask vector $\boldsymbol{m} = [m_1 \cdots m_d]^T$ is sampled from a Bernoulli distribution with probability $p$

$$\boldsymbol{m} \sim Bernoulli(p). \quad (4)$$

In practice the random mask $\boldsymbol{m}$ is scaled with $1/p$ so that no changes are needed during the testing phase of the model.

The random mask vector $\boldsymbol{m}$ is multiplied with the inputs (which are the outputs of the neurons in the previous layer) and the output is calculated as

$$\widetilde{\boldsymbol{a}}^{l-1} = \boldsymbol{a}^{l-1} \odot \frac{\boldsymbol{m}}{p}, \quad (5)$$

$$\boldsymbol{a}^l = \sigma\Big(\boldsymbol{W}^l \widetilde{\boldsymbol{a}}^{l-1} + \boldsymbol{b}^l\Big), \quad (6)$$

where $\odot$ is the elementwise product. In terms of weight perturbation, Dropout either turns off or scales all the outgoing weights from a neuron as follows

$$\widetilde{W}_{:,j} = \begin{cases} \boldsymbol{0} & \text{if } m_j = 0, \\ \frac{1}{p} W_{:,j} & \text{if } m_j = 1. \end{cases} \quad (7)$$

Randomly dropping neurons in the network forces individual neurons to learn useful representations on their own rather than developing dependencies on other neurons. During testing the weights are scaled with $p$, emulating the effect of averaging an ensemble of many ($2^{|\boldsymbol{m}|}$) models. Each model in the ensemble differs from the others by having different units dropped. The individual models in the ensemble are trained using a few training examples (same binary mask generated a few times) or none at all. Such an ensemble of models is interpreted as an approximation to the Bayesian model averaging [24].

In expectation, Dropout has been shown to be equivalent to penalizing the weights with $L_2$ norm for the cases of linear regression [30] and GLMs [32].

#### 2.3.2 Shakeout

Shakeout [15] is an extension of Dropout that results in both $L_1$ and $L_2$ regularization. Similar to Dropout, a Bernoulli random mask $\boldsymbol{m}$ with probability $p$ is generated, but the Shakeout operation perturbs the weights as follows:

$$\widetilde{W}_{ij} = \begin{cases} -c\,\text{sgn}(W_{ij}) & \text{if } m_j = 0, \\ \frac{1}{p} W_{ij} + c(\frac{1}{1-p})\text{sgn}(W_{ij}) & \text{if } m_j = 1, \end{cases} \quad (8)$$

where $c$ is the strength of $L_1$ regularization and sgn is the sign function. Thus, rather than zeroing out weights, Shakeout sets them to a constant $c$ with the opposite sign of the weight if the mask is zero and adds the constant $c$ to the weight if the mask is one.

#### 2.3.3 Dropout Variants

In addition to Shakeout, many variants of Dropout for feedforward neural networks have been proposed: Dropconnect [33] removes certain weights instead of complete units from the network; Alternating Dropout, where neurons that are retained in the current iteration are made more likely to be dropped in the next iteration; Standout [1] trains a separate network along with the main neural network that predicts an adaptive Dropout rate $p$; Monte-Carlo Dropout [8], where instead of scaling the weights to achieve averaging, multiple stochastic passes of the network are used to estimate the average, which gives a measure of the uncertainty in the prediction of the network; Swapout [28] samples network models from a much larger set of architectures, where neurons in each layer can be dropped, entire layers can be skipped or a combination of the two can be performed.

Another approach to learn robust network weights is variational learning [11, 3], where rather than learning a single value for each connection in the network, a probability distribution over each connection in the network is learned. If the distributions are modeled as Gaussian, these methods at least double the parameters in the network while having performance approaching that of Dropout.

Most of the aforementioned variants of Dropout are empirically motivated and do not have rigorous theoretical equivalence to deterministic regularization and model selection. To the extent of our knowledge, there is no stochastic regularization technique that is equivalent to a general $L_q$ penalty on the network weights.

## 3. Bridgeout

In this paper we propose the Bridgeout stochastic regularization, which is equivalent to an $L_q$ penalty on model weights. During training, a Bernoulli random mask matrix $M$ is generated with probability $p$. The Bridgeout operation then perturbs the weights as follows

$$\widetilde{W}^l = W^l + |W^l|^{\circ \frac{q}{2}} \odot \left(\frac{M}{p} - 1\right), \quad (9)$$

where $\circ$ is the elementwise power, $p$ is the hyperparameter determining the strength of regularization and $q$ is the hyperparameter determining the power of the norm.

Both the hyper-parameters are theoretically-grounded and have intuitive meanings: $q$ specifies the normed space from which model weights are learned and $p$ is the magnitude of the Lagrangian enforcing the normed space constraint. Normed spaces with $q < 2$ exhibit sparsity, which is practically desirable for faster convergence and reduced computational cost through network pruning.

Bridgeout subtracts the *normed* weight from the weight if the mask is 0 otherwise it adds a scaled *normed* weight as given below

$$\widetilde{W}_{ij}^l = \begin{cases} W_{ij}^l - |W_{ij}^l|^{\frac{q}{2}} & \text{if } M_{ij} = 0, \\ W_{ij}^l + |W_{ij}^l|^{\frac{q}{2}} \left(\frac{1-p}{p}\right) & \text{if } M_{ij} = 1. \end{cases} \quad (10)$$

The output of the $l$-th layer is then calculated as

$$\boldsymbol{\nu}^l = \widetilde{W}^l \boldsymbol{a}^{l-1} + \boldsymbol{b}^l, \quad (11)$$

$$\boldsymbol{a}^l = \sigma(\boldsymbol{\nu}^l). \quad (12)$$

To compute the gradient of the cost function for updating the network weights, the gradient of the pre-activations with respect to inputs and weights are given as

$$\frac{\partial \nu_i^l}{\partial a_j^{l-1}} = W_{ij}^l + |W_{ij}^l|^{\frac{q}{2}} \left(\frac{M_{ij}}{p} - 1\right), \quad (13)$$

$$\frac{\partial \nu_i^l}{\partial W_{ij}^l} = a_j^{l-1} \left(1 + \frac{q}{2}|W_{ij}^l|^{\frac{q}{2}-1} \left(\frac{M_{ij}}{p} - 1\right) \operatorname{sgn}(W_{ij}^l)\right), \quad (14)$$

respectively.

As indicated by Srivastava et al. [30], stochastic regularization with a high learning rate can cause weights to diverge. To help with convergence, we use the max-norm regularization [29] where each weight is constrained to be less than a threshold $|w| < t$ where $t$ is a hyperparameter. We set $t = 3.5$ in our experiments unless otherwise specified.

### 3.1. Equivalence to Bridge Regularization

**Theorem 1.** *For generalized linear models, the Bridgeout operation is equivalent to an $L_q$ penalty on the model weights.*

*Proof.* A generalized linear model (GLM), with parameter vector $\boldsymbol{\beta}$ and identity link function is given by

$$p_{\boldsymbol{\beta}}(y|x) = h(y)e^{(y\boldsymbol{x}\cdot\boldsymbol{\beta} - A(\boldsymbol{x}\cdot\boldsymbol{\beta}))}, \quad (15)$$

where $\boldsymbol{x}$ and $y$ are the input and response variables, $A$ is the log-partition function and $h$ is a function of the response variable $y$ [25]. Assume that the Bernoulli random mask $\boldsymbol{m}$ is scaled with $\frac{1}{p}$, then the Bridgeout weight perturbation can be expressed as feature noise as following

$$\widetilde{x}_j = x_j\big[1 + |\beta_j|^{\frac{q-2}{2}} \operatorname{sgn}(\beta_j)(m_j - 1)\big]. \quad (16)$$

With feature noise, the GLM is trained by minimizing the noise-marginalized negative log likelihood loss function over a dataset with $n$ samples as follows

$$\widehat{\boldsymbol{\beta}} = \underset{\boldsymbol{\beta} \in \mathcal{R}^d}{\operatorname{argmin}} \sum_{i=1}^n E_m[l_{\widetilde{\boldsymbol{x}},y}(\boldsymbol{\beta})], \quad (17)$$

where the loss function $l_{\widetilde{\boldsymbol{x}},y}$ can be split into two terms: the negative log likelihood term and a regularization term as follows

$$\sum_{i=1}^n E_m[l_{\widetilde{\boldsymbol{x}},y}(\boldsymbol{\beta})] = \sum_{i=1}^n l_{\boldsymbol{x},y}(\boldsymbol{\beta}) + R(\boldsymbol{\beta}), \quad (18)$$

where $R(\boldsymbol{\beta})$ is given by

$$R(\boldsymbol{\beta}) = \sum_{i=1}^n E_m[A(\widetilde{\boldsymbol{x}}^{(i)} \cdot \boldsymbol{\beta})] - A(\boldsymbol{x}^{(i)} \cdot \boldsymbol{\beta}). \quad (19)$$

In general the form of $R(\boldsymbol{\beta})$ is unknown, however, Wager et al. [32] have shown that a quadratic approximation provides good fidelity to $R(\boldsymbol{\beta})$. To get a quadratic approximation, we expand $A(\widetilde{\boldsymbol{x}}_i \cdot \boldsymbol{\beta})$ using Taylor series around $\boldsymbol{x}^{(i)} \cdot \boldsymbol{\beta}$

$$\widehat{R}(\boldsymbol{\beta}) = \sum_{i=1}^{n} \frac{A''(\boldsymbol{x}^{(i)} \cdot \boldsymbol{\beta})}{2} Var[\widetilde{\boldsymbol{x}}^{(i)} \cdot \boldsymbol{\beta}], \qquad (20)$$

where

$$Var[\widetilde{\boldsymbol{x}}^{(i)} \cdot \boldsymbol{\beta}] = E[(\widetilde{\boldsymbol{x}}^{(i)} \cdot \boldsymbol{\beta})^2] - E[(\widetilde{\boldsymbol{x}}^{(i)} \cdot \boldsymbol{\beta})]^2. \qquad (21)$$

The $E[(\widetilde{x}_j^{(i)} \beta_j)] = x_j^{(i)} \beta_j$ since the noise has unit expectation. We have $E[m_j^2] = \frac{1}{p}$ and $E[m_j] = 1$ thus

$$Var[\widetilde{\boldsymbol{x}}^{(i)} \cdot \boldsymbol{\beta}] = \sum_{j=1}^{d} \frac{1-p}{p} |\beta_j|^q \left(x_j^{(i)}\right)^2. \qquad (22)$$

Now by substituting the variance in the quadratic approximation of the regularizer, we have

$$\widehat{R}(\boldsymbol{\beta}) = \frac{1-p}{2p} ||\Gamma \boldsymbol{\beta}||_q^q, \qquad (23)$$

where

$$\Gamma = [X^T D X]^{\circ \frac{1}{q}}, \qquad (24)$$

$D$ is a diagonal matrix with elements $A''(\boldsymbol{x}^{(i)} \cdot \boldsymbol{\beta})$. ∎

**Corollary 1.1.** *For $q = 2$ the Bridgeout operation is equivalent to Dropout regularization for GLMs.*

*Proof.* Setting $q = 2$ in Equation 23, it becomes identical to the Dropout formulation given by Equation 11 in Wager et al. [32]. ∎

## 4. Experimental Results

In this section, we provide experimental results to show the sparsity inducing property of Bridgeout and its effectiveness in the case of synthetic data classification with noisy features. We also evaluate Bridgeout for image classification using MNIST and Fashion-MNIST datasets.

### 4.1. Characterizing Bridgeout

#### 4.1.1 Sparse Weight Distribution

In order to demonstrate the effect of Bridgeout regularization on model weights, we apply Bridgeout regularization to a linear regression problem with synthetic data. The data was generated as follows: 400 100-dimensional samples were generated from a Gaussian distribution, a Gaussian random weight matrix of dimensions $100 \times 10$ was used to transform the input samples to 10-dimensional output samples. A linear regression model with different regularization methods was trained for 5000 iterations using gradient descent. The normalized histograms of the weight matrices (Figure 1) illustrate that a smaller value of $q$ in Bridgeout results in weight distribution concentrated around zero. As expected, setting $q = 2$ in Bridgeout results in the same weight distribution as Dropout.

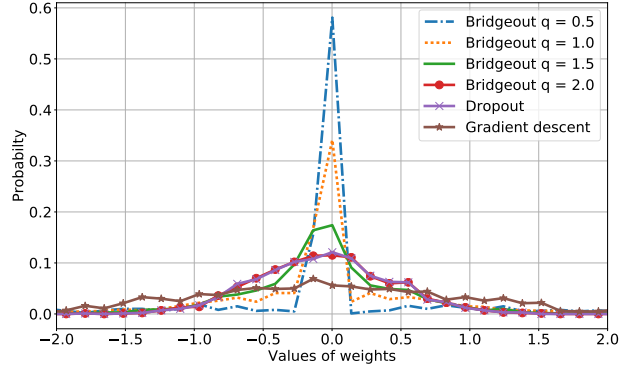

Figure 1: Distribution of weights of the linear regression model trained with stochastic regularization techniques

To see the impact of Bridgeout regularization on the weights of non-linear models, we used an autoencoder consisting of $728 - 256 - 728$ neurons. We used the MNIST dataset [19] to train the autoencoder for 500 epochs using backpropagation. Different regularizations were applied to the encoder part of the autoencoder. As in the case of linear regression, Bridgeout with smaller values of $q$ resulted in sparser weight distributions as shown in Figure 2.

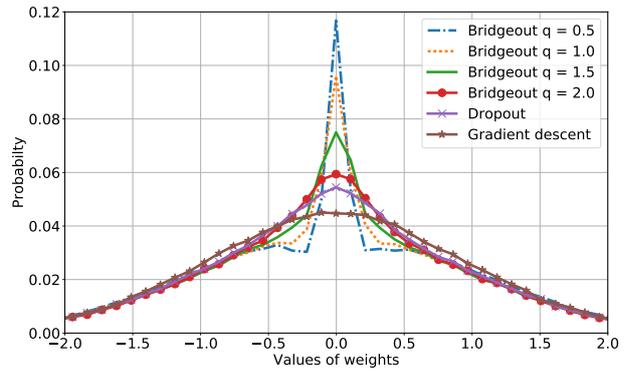

Figure 2: Distribution of the weights of the encoder of the autoencoder trained with stochastic regularization.

Visualizing the weights of the encoder in Figure 3, we see that both with Bridgeout and Dropout each neuron in the network learns interesting features by itself while in the case of standard backpropagation individual neurons do not seem to have learned any specific features indicating dependencies on other neurons, resulting in a fragile network.

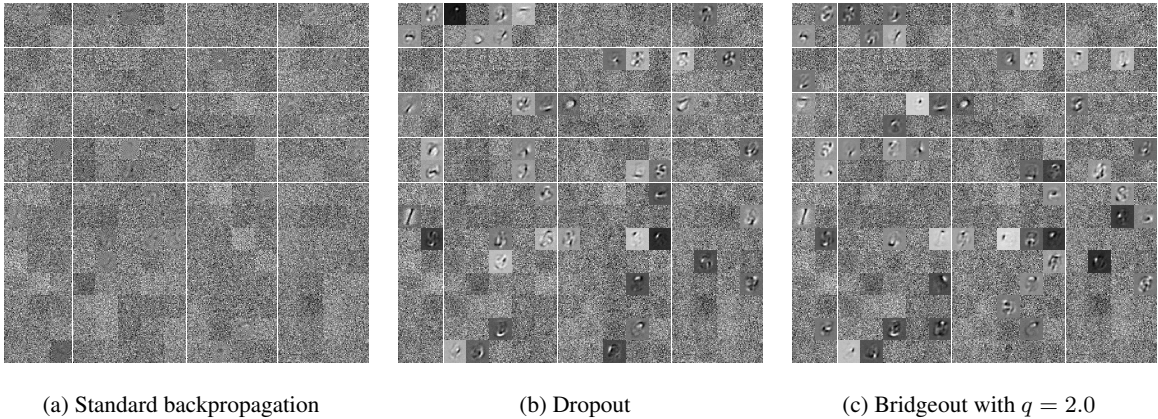

(a) Standard backpropagation  (b) Dropout  (c) Bridgeout with $q = 2.0$

Figure 3: Visualization of the weights learned by the encoder neural network trained with different regularization methods

### 4.1.2 Synthetic Data Classification

For classification tasks with many noisy predictors, we expect that a regularization norm other than $q = 2$ will provide better performance. To show the impact of Bridgeout in the case of noisy predictors, we adopt the experimental setup proposed by Liu et al. [21]. For each trial of the experiment we generate 400 samples from $\{0, 1\}^{20}$ uniformly. For each input sample the class label $y$ is assigned as $\text{sgn}(f(x))$, where

$$f(x) = 2x_0 + 4x_1 + 4x_2 - 4.8. \qquad (25)$$

Thus only the first three predictors in the input are important for the class labels while the other 17 are noise variables. Based on this we expect that $L_2$ will not be a good regularizer in this case. For each experiment a test set of 3000 was generated. A learning rate of 0.001 with 8000 iterations of gradient descent optimizer was used. Retention probability $p$ was set to 0.5, while for Bridgeout the norm $q$ was set to 1.0 and for Shakeout the $L_1$ regularization strength was set to 0.3. The experiments were repeated 50 times and the mean and standard error of the misclassification rate are reported in Table 1. As can be seen from the results Dropout performs poorly because it spreads out the weights and forces them to be non-zero, effectively expanding the search space from 3-dimensions to 20-dimensions. On the other hand, Bridgeout and Shakeout result in the best performance due to their sparsity inducing nature.

Table 1: Binary classification with logistic regression

| Method | Test Error % |
| --- | --- |
| Gradient Descent | $0.279 \pm 0.058$ |
| Dropout | $1.282 \pm 0.165$ |
| Shakeout | $\mathbf{0.054 \pm 0.011}$ |
| Bridgeout | $\mathbf{0.047 \pm 0.038}$ |

### 4.2. Image Classification

We evaluated the performance of Bridgeout in comparison to Dropout and Shakeout on standard image classification datasets. For all our experiments we used the Adam [16] optimizer with all the default values that are highly optimized to Dropout. We initialized the weights using Xavier initialization [9] for all the layers. For hyper-parameter optimization we used the Tree-structured Parzen Estimator (TPE) algorithm [2] with 30 evaluations for all the methods. For Dropout we optimized the retention probability $p$, for Shakeout we optimized the retention probability $p$ and $L_1$ regularization strength $c$, and for Bridgeout we optimized the retention probability $p$ and the norm $q$.

In all experiments we used the training set for training the models, the validation set to select hyperparameters and the test set only for reporting the error rate of the trained models. Once the hyperparameters were obtained, 5 independent networks with different random seeds were trained. The mean and standard error of the misclassification rate was reported for each method.

### 4.2.1 Classifying MNIST

MNIST is a benchmark dataset for image classification tasks consisting of grayscale images of handwritten digits from 0 to 9 of size $28 \times 28$ [19]. MNIST consists of 50000 training images, 10000 validation images and 10000 test set images. To check classification performance and the behaviour of gradients, while keeping the training time to a minimum for hyper-parameter optimization, we trained deep neural networks with three fully connected layers of size 200 with non-linear activations, followed by a softmax output layer of size 10. Two different non-linear activations were used for the hidden neurons in the network: sigmoid and rectified linear units (ReLU). We applied regularization to all the three fully connected layers. No preprocessing was applied to the MNIST dataset except normalizing the pixel values to $[0, 1]$. We trained the network with subsets of the training set to see the impact of overfitting and used the full validation set to select the hyperparameters.

After hyperparameter optimization of $p$ and $q$, we found that the optimal value of the norm $q$ for Bridgeout varied across different subsets of the dataset (Table 2) demonstrating that $q \in \{1, 2\}$ are not the optimal values for regularization. The error rates for the MNIST test set (Table 3) show that for this task Bridgeout resulted in the lowest error rates across all training set sizes. Moreover, as shown in Figure 4, when sigmoid activations are used, Bridgeout results in larger gradients in the near-input layers when sigmoid units are used. This could help in avoiding the gradient vanishing problem that exist in networks with sigmoid activations.

Table 2: The optimal norm $q$ in Bridgeout varies for different sampling of the dataset and is not restricted to $\{1, 2\}$.

| Training set size | 3K | 5K | 8K | 20K | 50K |
|---|---|---|---|---|---|
| DNN-Sigmoid-MNIST | 0.99 | 1.13 | 1.23 | 0.89 | 1.07 |
| DNN-ReLU-MNIST | 1.68 | 1.36 | 0.85 | 1.27 | 1.76 |
| CNN-MNIST | - | 0.75 | - | - | 0.84 |
| CNN-FMNIST | - | 0.66 | - | - | 1.54 |

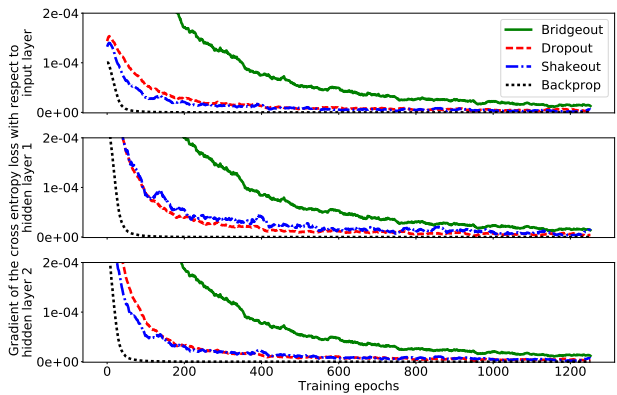

Figure 4: Average gradients of the cost function calculated as $\frac{1}{|W^l|} \sum_{w \in W^l} \frac{\partial J}{\partial w}$ of the sigmoid deep neural network trained with a subset of MNIST of size 5000.

We also trained a convolutional neural network (CNN) with the architecture similar to the one used by Wan et al. [33]. The CNN consisted of two convolutional layers with 32 and 64 channels of filter size $5 \times 5$, each with max pooling of size $2 \times 2$ and ReLU activation. The convolutional layers were followed by a fully connected layer of size 150 with ReLU activation. Regularization was applied to the fully connected layer. Finally a fully connected layer of size 10 with softmax activation was used to output the image label probabilities.

Similar to the deep neural network case, for CNNs, a non-integral value was found to be optimal for the norm $q$ in Bridgeout. Bridgeout resulted in the lowest classification errors for both the full MNIST and a subset of MNIST dataset as shown in Table 5. As shown in Figure 5(left), Bridgeout results in higher gradients of the cost function specifically with respect to the input convolutional layer. Compared to the other methods, Bridgeout takes longer to converge but results in the lowest validation error as shown in Figure 5(right).

### 4.2.2 Classifying Fashion-MNIST

Fashion-MNIST [35] is a new dataset that is very similar in structure and size to MNIST but comprises of images of fashion products instead of handwritten digits. Fashion-MNIST consists of images belonging to 10 classes of fashion products such as t-shirts, trousers and bags etc. as shown in Figure 6. Thus Fashion-MNIST provides a semantically more challenging alternative to MNIST.

We used the same convolutional neural network architecture for Fashion-MNIST as used for the MNIST dataset described in the previous section. Table 6 shows the results of training the CNN with 5000 and 50000 training images from Fashion-MNIST. For the training set of size 5000, Bridgeout resulted in around $1\%$ improvement over Dropout while for the full training set, Bridgeout and Dropout resulted in similar performance. This indicates that Bridgeout can effectively reduce overfitting when the dataset size is comparatively small.

## 5. Practical recommendation for hyperparameters.

For a particular problem it is recommended to optimize for $p$ over $[0.3, 0.7]$ and for $q$ over $[0.5, 2.0]$. In our experiments, we found that setting $p = 0.5$ and optimizing for $q$, the optimal value of $q$ reaches 2.0 as the dataset size increases as shown in Table 7. Depending on the problem at hand, $q$ can be decreased to increase regularization strength and sparsity of the weights.

## 6. Discussion

Dropout and other stochastic regularization techniques are often used to reduce overfitting in image classification with deep neural networks. Many problems, including image classification, can benefit from a sparsity inducing penalty while keeping the properties of stochastic regularization. Shakeout augments Dropout by adding an $L_1$ norm term to encourage weight sparsity. Bridgeout, on the other hand, allows for a fractional norm that can be tuned to better match the shape of the penalty to the problem at hand. Im-age classification experiments with Bridgeout did yield optimal values of $q$ less than 2, which encourages sparsity, and resulted in the best performances on image classification using both fully connected and convolutional neural networks.

Table 3: Error rates (%) of deep neural network with sigmoid activations, trained on MNIST dataset with different training set sizes.

| Training set size | 3000 | 5000 | 8000 | 20000 | 50000 |
|---|---|---|---|---|---|
| Backprop | 8.586 ± 0.064 | 6.276 ± 0.056 | 4.688 ± 0.020 | 3.136 ± 0.024 | 2.010 ± 0.013 |
| Dropout | 7.752 ± 0.127 | 5.508 ± 0.037 | 4.362 ± 0.036 | 2.760 ± 0.025 | 1.858 ± 0.045 |
| Shakeout | 8.430 ± 0.106 | 6.594 ± 0.088 | 5.112 ± 0.075 | 3.198 ± 0.044 | 1.960 ± 0.021 |
| Bridgeout | **6.484 ± 0.031** | **4.676 ± 0.044** | **3.752 ± 0.049** | **2.470 ± 0.032** | **1.642 ± 0.028** |

Table 4: Error rates (%) of deep neural network with ReLU activations, trained on MNIST dataset with different training set sizes.

| Training set size | 3000 | 5000 | 8000 | 20000 | 50000 |
|---|---|---|---|---|---|
| Backprop | 7.707 ± 0.080 | 5.884 ± 0.066 | 4.662 ± 0.068 | 2.886 ± 0.034 | **1.646 ± 0.046** |
| Dropout | 6.656 ± 0.133 | 4.788 ± 0.069 | 3.880 ± 0.112 | 2.512 ± 0.041 | 1.716 ± 0.018 |
| Shakeout | 6.782 ± 0.077 | 5.046 ± 0.071 | 4.006 ± 0.064 | 2.712 ± 0.040 | 1.708 ± 0.044 |
| Bridgeout | **5.974 ± 0.055** | **4.442 ± 0.059** | **3.626 ± 0.056** | **2.370 ± 0.016** | 1.612 ± 0.061 |

Table 5: Error rates (%) of convolutional neural network trained on MNIST dataset.

| Training set size | 5000 | 50000 |
|---|---|---|
| Backprop | 2.972 ± 0.064 | 0.808 ± 0.015 |
| Dropout | 1.942 ± 0.058 | 0.638 ± 0.016 |
| Shakeout | 1.944 ± 0.057 | 0.628 ± 0.023 |
| Bridgeout | **1.846 ± 0.016** | **0.600 ± 0.017** |

Table 6: Error rates (%) of convolutional neural network trained on Fashion-MNIST dataset.

| Training set size | 5000 | 50000 |
|---|---|---|
| Backprop | 13.012 ± 0.086 | 8.718 ± 0.084 |
| Dropout | 12.054 ± 0.088 | **7.724 ± 0.077** |
| Shakeout | 11.862 ± 0.127 | 7.898 ± 0.095 |
| Bridgeout | **11.152 ± 0.071** | 7.614 ± 0.074 |

Table 7: Optimal $q$ for $p = 0.5$ for Bridgeout applied to DNN trained on the MNIST dataset.

| Training set size | 3K | 5K | 8K | 20K | 50K |
|---|---|---|---|---|---|
| Optimal $q$ | 1.57 | 1.63 | 1.91 | 1.99 | 1.97 |

Both Dropout and Bridgeout resulted in interesting learned features in individual neurons of the network, as indicated in Figure 3; however, they did so in very different ways. Neurons in the networks trained with Dropout are forced to learn representations that are useful in the absence of other neurons since during training only a fraction $p$ of the neurons are present. Bridgeout, on the other hand, forces neurons to learn robust representations in a more adversarial environment where synapses are randomly damped-down (a norm of the weight is subtracted) or spiked-up (a scaled norm is added) as evident from Equation 10. This could be biologically more plausible since activities of the neurons in the brain are noisy.

Since Bridgeout does not zero out weights during training, it prevents the vanishing gradients problem that is common in Dropout-based regularization. Better gradients are important for training very deep neural networks as was shown by He et al. [13] with the residual neural networks. Bridgeout could potentially help in training deep networks in a manner similar to the residual learning paradigm, which we plan to investigate in future studies with deeper networks than used in the present study.

It is interesting to note that the Shakeout perturbation becomes analytically equivalent to the Dropout perturbation when the $L_1$ regularization strength $c$ is set to zero. On the other hand, for the norm $q = 2$, the Bridgeout weight perturbation (Equation 10) is analytically different from the Dropout perturbation (Equation 7), but, in expectation, they are equivalent, as shown theoretically in Corollary 1.1 and demonstrated empirically in Figure 1.

Regularization techniques work well in practice and result in superior classification performance. The improvement in performance due to the stochastic regularization techniques is high, specifically in the scarce training data regime. As shown in Table 3, for training set sizes less than 20000 Bridgeout results in about 2% improvement in the generalization error over standard backpropagation. However, recently Zhang et al. [38] showed that deep neural networks with or without regularization have sufficient capacity to achieve zero training error on image classification where the labels are assigned randomly. Thus, it is still an open question as to why neural networks generalize better even though they have much higher capacity than the one required for the task. Nevertheless, regularization remains standard practice and Bridgeout could be used in many current real-world problems, such as biomedical image classification/diagnosis where labelled data is limited.

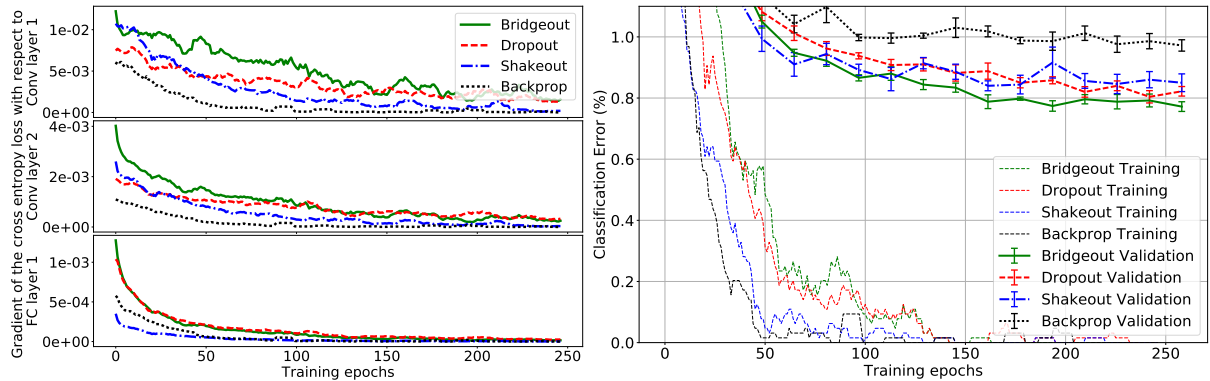

Figure 5: CNN trained with MNIST, average gradients of the cost function with respect to different layers (left), classification errors (right)

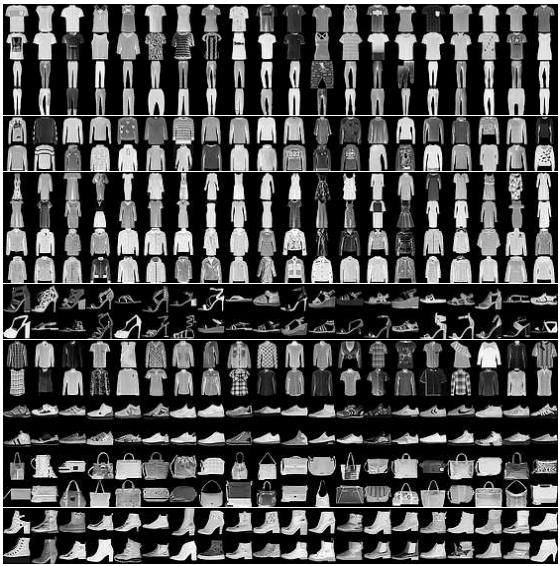

Figure 6: Fashion-MNIST [35] dataset comprising of images of fashion products.

In order to provide a fair comparison with respect to hyperparameter optimization, we chose the relatively simple 4-layer neural network and a 4-layer CNN, with relatively easy datasets MNIST and Fashion-MNIST. Besides being simple, MNIST is also relatively easy to generalize from very small training sets, thus achieving better performance on these datasets with regularization is challenging. Also, the use of simple models makes improvement over backpropagation challenging since there is relatively less overfitting. We expect the benefits to Bridgeout to be greater for larger architectures or problems with scarce data where regularization is more important to combat overfitting. As future work, we plan to test Bridgeout with more complex architectures and datasets such as CIFAR-100 and ImageNet.

## 7. Summary

In this paper, we have presented Bridgeout: the first stochastic regularization technique that is equivalent to an $L_q$ penalty on the model weights. We proved theoretically and empirically that Dropout is a special case of Bridgeout. Evaluation on image classification tasks using neural networks showed that the flexibility and sparsity-inducing properties of Bridgeout outperform Dropout and Shakeout in terms of classification accuracy.

## 8. Acknowledgement

This work was supported in part by the Natural Sciences and Engineering Research Council of Canada (NSERC).